%% file: acl_latex.tex
\pdfoutput=1

\PassOptionsToPackage{table}{xcolor}

\documentclass[11pt]{article}

\usepackage[]{acl}

\usepackage{times}
\usepackage{latexsym}

\usepackage[T1]{fontenc}

\usepackage[utf8]{inputenc}

\usepackage{microtype}

\usepackage{inconsolata}

\usepackage[table]{xcolor}
\usepackage{graphicx}
\usepackage{floatrow}
\usepackage{subcaption}
\usepackage{multirow}
\usepackage{amssymb}
\usepackage[flushleft]{threeparttable}
\usepackage{amsmath}
\usepackage{booktabs}

\usepackage{lipsum}
\usepackage{pifont}
\usepackage[normalem]{ulem}
\definecolor{applegreen}{rgb}{0.0, 0.5, 0.0}
\definecolor{lightblue}{HTML}{D2F9F9}

\title{Generative Deduplication For Socia Media Data Selection}

\author{
    Xianming Li $^{1}$, 
    Jing Li $^{1, 2}$\thanks{\ \ Corresponding author} \\
    $^{1}$ Department of Computing \\
    $^{2}$ Research Centre on Data Science \& Artificial Intelligence \\
    The Hong Kong Polytechnic University
    \\
  \texttt{xianming.li@connect.polyu.hk}, 
  \texttt{jing-amelia.li@polyu.edu.hk}\\
 }

\begin{document}
\maketitle
\begin{abstract}
Social media data exhibits severe \textit{redundancy} caused by its noisy nature.
It leads to increased training time and model bias in its processing. 
To address this issue, we propose a novel Generative Deduplication framework for social media data selection by removing semantically duplicate data.
While related work involves data selection in task-specific training, our model acts as an efficient pre-processing method to universally enhance social media NLP pipelines.
Specifically, we train a generative model via self-supervised learning to predict a keyword to capture the semantics of noisy social media text for deduplication. 
Meanwhile, time-dimensional Gaussian noise is added to improve training complexity and avoid learning trivial features.
Extensive experiments suggest that our model can better reduce training samples while improving performance than baselines. 
The results show our model's potential to broadly advance social media language understanding in effectiveness and efficiency. \footnote{Code is available at: \url{https://github.com/4AI/generative_deduplication}.}

\end{abstract}

\section{Introduction}
\input{introduction}

\section{Related Work}
\input{related_work}

\section{Generative Deduplication}
\input{method}

\section{Experimental Setup}
\input{setup}

\section{Experimental Results}
\label{sec::main-experiment}
\input{main_experiment}

\section{Conclusion}
In this paper, we have introduced a novel pre-processing method called generative deduplication for social media data selection. 
It tackles the semantic redundancy bias in noisy social media data. 
Extensive experiments have suggested that generative deduplication can significantly reduce the training cost of a model (in time and resources) while improving social media language understanding. 


\section*{Ethics Statement}
In our empirical study, we use publicly available social media understanding datasets that have been widely used in previous studies. These datasets typically do not have direct societal consequences. Our model introduces a novel paradigm, generative deduplication, for social media data selection. 
The proposed generative deduplication reduces the number of training samples, resulting in decreased computational resources, which is beneficial for the environment.

\section*{Limitations}
We have initially tested the proposed generative deduplication method on widely used social media data, specifically TweetEval. In the future, we plan to extend our evaluation to additional social media datasets, such as Reddit TIFU \citep{kim2018abstractive} and GoEmotions \citep{demszky2020goemotions}. Furthermore, generative deduplication is a general technique that can be applied to different contexts beyond social media. We will explore its applicability in broader scenarios.

\section*{Acknowledgements}
\input{acknowledgement}

\bibliography{anthology,custom}

\appendix

\end{document}

%% file: introduction.tex
Social media is an abundant resource with vast real-time user-generated content, providing valuable insights into the world and society. 
It has
benefited various applications, such as stance detection \citep{glandt-etal-2021-stance} and content recommendation \citep{zeng-etal-2020-dynamic}, taken advantage of cutting-edge NLP practices. 
However, a common challenge NLP models may face is the severe \textit{redundancy} of social media data \citep{tao2013groundhog} caused by its noisy nature \citep{zhang-etal-2023-vibe}. Here, we define \textit{redundancy} as semantically similar content that leads to information overload and model bias.

\input{figure_example}

The redundant data not only increases the training cost of a model (in time and resources) but also results in the \textbf{redundancy bias} adversely affecting its performance \citep{lee-etal-2022-deduplicating}.
To illustrate this point, we show some TweetEval examples in Figure \ref{figure_example}.
The duplicated tweets containing ``pumpkin'' and ``neutral'' sentiment labels bias the model to connect ``pumpkin'' to ``neutral wrongly'' sentiment, meanwhile rendering unnecessary training costs.

To tackle the redundancy problem, we explore the solution from \textit{data selection} \citep{liu-etal-2019-reinforced, paul2021deep, lee-etal-2022-deduplicating, xie2023data}. 
The goal is to select a subset of relevant data from a larger dataset to benefit the model training. 
It has profoundly affected the performance of large language model pretraining \citep{xie2023data} and its downstream task finetuning \citep{yu-etal-2023-cold}.

In data selection, most work focuses on how to find useful data via augmentation or retrieval for ``data addition'' \citep{axelrod-etal-2011-domain, ruder2017data, liu-etal-2019-reinforced, xie2023data}. 
In contrast, others center on ``\textbf{data deduplication}''  to remove semantically duplicated data and show its positive effects in pretraining
\citep{lee-etal-2022-deduplicating}.

Our work aligns with the ``deduplication'' line.
Previous work has shown its help in social media NLP, such as shingles \citep{broder1997resemblance} and simhash \citep{manku2007detecting}. 
However, these efforts rely on surface linguistic features rather than semantics.
While textual semantic similarity models \citep{sbert-nils-2019,simcse_gao_2021,li2023angle} can be easily applied to the deduplication, the pairwise comparison would render high complexity. 
Moreover, our work differs from existing efforts \citep{xia2024less} to engage data selection in 
task-specific training. 
Instead, we aim to provide a pre-processing method for social media data, which can be easily plugged into various NLP pipelines.



To that end, we propose a novel semantic deduplication approach, \textbf{Generative Deduplication}, for social media data selection.
Specifically, we adopt a generative model
as the generative backbone and train it with a self-supervised task to generate a
keyword from the input text. 
Here, we train the generative backbone 
for only one epoch.
Duplicate text undergoes multiple optimizations, enabling more accurate keyword prediction than non-duplicate text with a single optimization.
Moreover, we improve the training difficulty with Time-dimensional Gaussian Noise (TGN) to prevent trivial feature learning in one epoch, limiting keyword prediction for non-duplicates.
Hence, we can consider samples with correct keyword prediction as duplications and remove them.
This way, we allow a computational complexity of $O(n)$, where $n$ is the data size, and avoid pairwise comparison.


To the best of our knowledge, \textit{we are the first to propose Generative Deduplication for social media data selection and study its broad impact on downstream social media language understanding.
}

In experiments, the deduplication experiment indicates that our model enables the best data quality with less training time.
Then, the results on the TweetEval benchmark show that our selected data allows performance gains with much shorter training time on varying downstream models and tasks.
For example,  for LLaMA on sentiment analysis, our model reduces the training set ($50.9\%$) and training time ($42.9\%$) yet improves the macro recall from $73.0$ to $73.5$. 
Next, the ablation study shows the positive contributions of varying modules. The general short text classification experiments suggest that our method also benefits general scenarios.
Lastly, we interpret our model's superiority with more analyses.

In summary, our contributions are as follows:

$\bullet$ We explore the redundancy issue in social media data and disclose its effects on biasing models.

$\bullet$ We propose a novel generative deduplication model to shortlist data and tackle redundancy bias.

$\bullet$ Extensive experiments reveal generative deduplication can help reduce redundancy and broadly improve social media language understanding.

%% file: figure_example.tex
\begin{figure}
    \centering
    \includegraphics[width=1.0\textwidth]{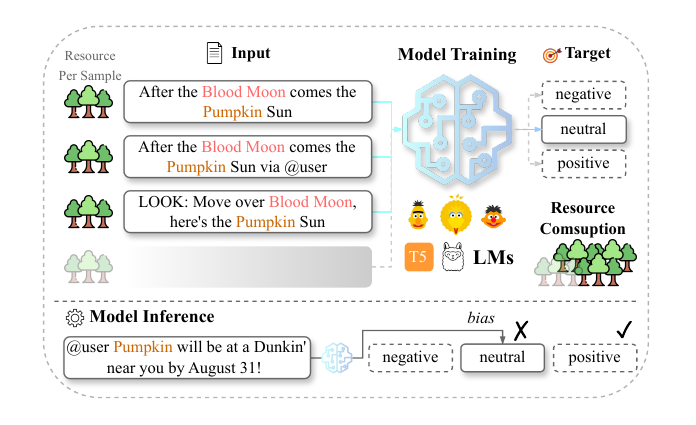}
    \caption{An redundancy example on TweetEval \citep{barbieri-etal-2020-tweeteval} sentiment analysis. 
    The green trees denote training costs.
    The duplicated tweets biased the models to connect ``pumpkin'' to ``neutral'' sentiment.
    }
\vspace{-1em}    \label{figure_example}
\end{figure}

%% file: related_work.tex
The proposed Generative Deduplication is in line with data selection. 
It is an essential technique for selecting helpful data to benefit the training of downstream tasks.
Many previous studies center on domain adaptation \citep{moore-lewis-2010-intelligent, feng-etal-2022-automatic, xie2023data}, where they  
selected data that aligns with the target distribution from vast data, aiming to improve performance in specific domains. 
More recently, \textit{deep learning} techniques \citep{Coleman2020Selection, mindermann2022prioritized} have been used for better data selection.

Given recent language model advances, some work, such as  \citep{yao2022nlp, schoch-etal-2023-data}, explored \textit{retrieval} and \textit{augmentation}  to ``select and add'' relevant data from external resources for task-specific training. 
However, prior work observed these practices may result in a significant amount of duplicate data, which introduces redundancy \citep{xie2023data} and adversely affects performance \citep{hernandez2022scaling}. 
To mitigate this redundancy issue, \textbf{deduplication}  \citep{tirumala2023d} can be applied to shortlist duplicate data to allow more effective and efficient training.

Our work aims to adopt deduplication to address redundancy bias  \citep{tao2013groundhog,zhang-etal-2023-vibe} in social media data. 
However, existing deduplication methods \citep{broder1997resemblance,manku2007detecting,hajishirzi2010adaptive} mainly rely on surface linguistic features, unable to handle
semantic-level duplication prevalent in noisy social media data. Meanwhile, directly applying models based on textual semantic similarity \citep{sbert-nils-2019,li2023angle,li-li-2024-bellm} may involve pairwise comparison and result in high deduplication cost. 
In addition, existing methods engage data selection in 
end-to-end 
task-specific training \citep{xia2024less}.
In contrast, our work focuses on data selection for pre-processing, which can be seamlessly integrated into various NLP pipelines.

%% file: method.tex
This section will elaborate on the proposed generative deduplication with an overall framework in Figure \ref{figure_framework}. 
We will first introduce the problem formulation in Section \ref{sec::problem_formulation}.
Then, we describe the generative training in Section \ref{sec::gen-training}, followed by 
the inference as the deduplication stage in Section \ref{sec::infer-as-dedup}.

\input{figure_framework}

\subsection{Problem Formulation}
\label{sec::problem_formulation}
Given a corpus $\mathcal{C}$ consisting of $n$ texts, $\{t_1, ..., t_n\}$, generative deduplication aims to identify semantically duplicate texts using generative models. 
The identified duplicate texts are then removed, forming a smaller corpus $\mathcal{SC}$ comprising $m$ texts, where $m \le n$.
To achieve this, two stages are involved: generative training and inference. 
We will discuss them in subsequent sections. 

\subsection{Generative Training}
\label{sec::gen-training}

The generative deduplication involves two important designs: generative self-supervised training to predict keywords and adding time-dimensional Gaussian noise to increase training difficulty.

\paragraph{Generative Self-supervised Training.}
We employ a novel self-supervised learning task of \textit{keyword generation} for social media texts. 
We are motivated by the noisy nature of social media data, and the inherent data sparsity can limit the explicit indicators of semantic similarity \citep{zeng-etal-2018-topic}. 
Keywords as condensed post-level representations can bridge the gap and enable better exploration of semantic similarity for deduplication purposes.

To implement this, we first apply the popular toolkit KeyBERT \citep{grootendorst2020keybert} for keyword extraction. 
Then, the extracted keyword serves as the target.
Specifically, the  contextual representations of text $t_j$ in $\mathcal{C}$ is obtained as follows: $\mathbf{H} = \mathrm{g}(t_j)$, 
where $\mathrm{g}(\cdot)$ represents the generative backbone.
For each training sample in $\mathcal{C}$, the objective is to minimize the sum of the negative likelihood of keyword tokens $\{k_1, ..., k_l\}$, where $l$ is the length of the keyword tokens, as follows:
\begin{equation}\small
    \mathcal{L}_{g} = - \sum^{l}_{i=1} \mathrm{log}\ \mathrm{g}_{\theta} (k_i | t_j;k_0, k_1, ..., k_{i-1}).
\end{equation}
Here, $\theta$ is the learnable parameters, $t_j$ is the $j$-th input text in $\mathcal{C}$, $k_0$ denotes the pre-defined start token, and $\mathrm{g}(\cdot)$ represents the generative backbone. 
Notably, the self-supervised training runs only one epoch for a sufficiently large prediction gap of duplicate and non-duplicated data (see Section \ref{sec::infer-as-dedup}).

\paragraph{TGN: Time-dimensional Gaussian Noise.}
\label{sec::tgn}

According to the scaling law \citep{kaplan2020scaling}, large generative models possess exceptional language understanding capabilities due to the large-scale pre-training.
However, this can negatively affect deduplication performance because even non-duplicate texts might have accurate keyword predictions in one-epoch training, hindering the separation of duplicate and non-duplicate data.

To address this concern, we propose a novel TGN to add noise and increase training difficulties.
It aims to avoid learning trivial features to limit the keyword prediction for non-duplicate data.  
It first generates binary masks $\mathbf{M}$ using the Bernoulli distribution, where each time step is assigned a value of $0$ or $1$ based on a given probability $p$:
\begin{equation}
\small
\mathbf{M} \sim \mathrm{Bernoulli}(p)
\end{equation}

Then, the time steps with a mask value of $1$ are selected, and their corresponding features are entirely replaced by Gaussian noise. 
This process is shown in Figure \ref{figure_framework}, and the equation as follows:
\begin{equation}\small
    \begin{split}  \widehat{\mathbf{H}} &= \mathbf{H} \odot (1 - \mathbf{M}) + \mathbf{M} \odot \mathbf{G} \\
    \mathbf{G} &\sim \mathcal{N}(\mu, \sigma^2),
    \end{split}
\end{equation}
where $\mathbf{G}$ denotes the standard Gaussian distribution, which has the mean $\mu$ and standard deviation $\sigma$.


\subsection{Inference as Deduplication}
\label{sec::infer-as-dedup}
After the one-epoch self-supervised training with TGN, the model makes inferences to predict keywords for deduplication. 
The trained generative backbone generates keywords for all texts in $\mathcal{C}$ using beam search during this stage. 
Specifically, for text $t_i$ in $\mathcal{C}$, its keyword is generated as follows:
\begin{equation}\small
    \widehat{K} = \mathrm{g}(t_i;b),
\end{equation}
where $b$ is the beam size for beam search.
Here, we consider the text duplicates if they can accurately replay the target keyword through the trained generative backbone.
Our intuition is that the model is more likely to comprehend semantically duplicate texts than non-duplicates because of the multiple optimizations to the former. 
Consequently, the duplicate text will result in higher chances for the model to replay the keywords after one-epoch training.
Based on this, we compare the generated keyword with the target keyword for deduplication:
\begin{equation}\small
    \begin{split}
        \mathrm{IsDup}(t) &= \left\{\begin{matrix}
         1 & if\ \widehat{K} = K \\ 
         0 & otherwise
        \end{matrix}\right. \\
    \end{split}
\end{equation}
$K$ is the target keyword for $t$, $\widehat{K}$ is its predicted keyword from the
generative backbone $\mathrm{g}(\cdot)$, and $1$ and $0$ indicate ``yes'' and ``no'' for deduplication.

%% file: figure_framework.tex
\begin{figure*}
    \centering
    \includegraphics[width=0.95\textwidth]{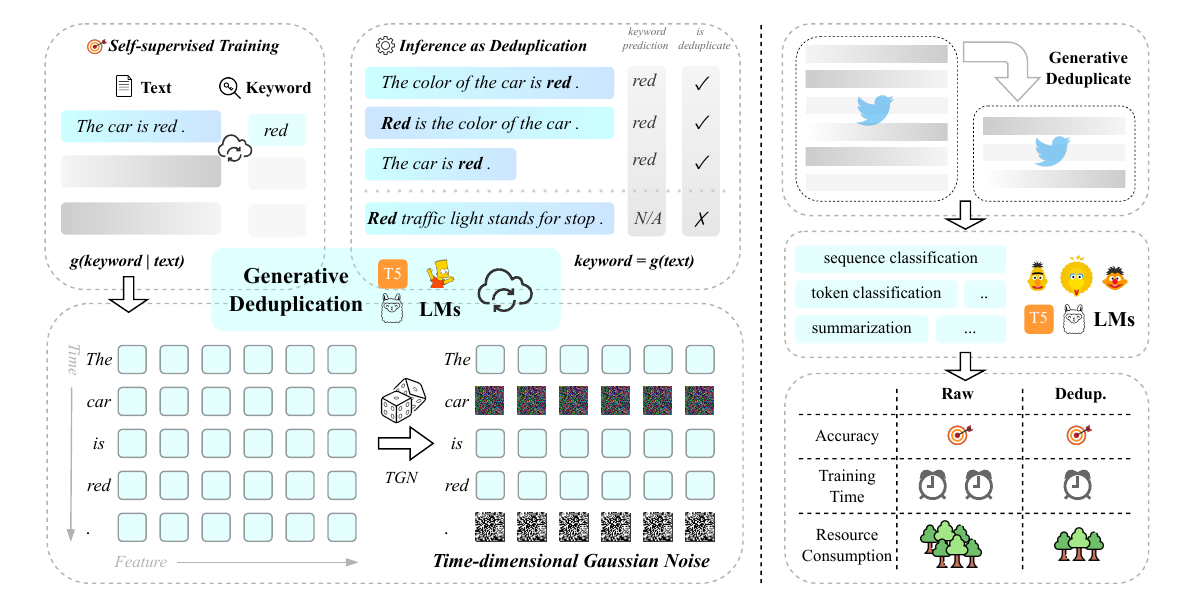}
    \caption{The generic framework of the proposed generative deduplication (GD). It includes two parts. 1) The left part shows the detailed GD process. In the \textit{self-supervised training stage}, the model encodes input text and learns to generate the corresponding keyword. The \textit{Time-dimensional Gaussian Noise (TGN)} will be applied to increase training difficulties and avoid trivial feature learning in the training stage. In the \textit{inference as deduplication stage}, text with correct keyword prediction is identified as a duplicate. 
    2) The right part depicts the downstream applications. First, the training set is deduplicated using GD. Then, the deduplicated training set is used for training and inference. By doing so, It can reduce training samples and resource consumption while improving accuracy.}
    \label{figure_framework}
\end{figure*}

%% file: setup.tex
\paragraph{Datasets.}
For the \textbf{deduplication experiment},  we use the MRPC dataset \citep{dolan-brockett-2005-automatically}, where texts are labeled as either \textit{equivalent} or \textit{non-equivalent} based on semantic duplication. For \textbf{social media language understanding},
we evaluate our model on 7 widely used Twitter datasets, including 6 tasks from TweetEval \citep{barbieri-etal-2020-tweeteval}: \underline{Emoji} Predication, \underline{Hate} Speech Detection, \underline{Offensive} Language Identification, \underline{Sentiment} Analysis, \underline{Stance} Detection, and \underline{Irony} Detection, and \underline{Humor} Detection from SemEval-2021 task 7 \citep{meaney-etal-2021-semeval}. The statistics of adopted datasets are listed in Table \ref{table_dataset_statistics}.
Due to the limitations of space and computational resources, we use the \underline{Emoji} dataset in the ablation study and discussion. It is more challenging (with 20 labels) and larger than other datasets, making it ideal for evaluating the performance of the proposed model.

\input{table_dataset_statistics}

\paragraph{Evaluation Metrics.} 
For \textbf{deduplication}, we report the F1 score for \textit{equivalent} (duplicate text), marked as F1$^{eq}$. We also report the deduplication time in seconds. 
For \textbf{social media language understanding}, we follow prior works \citep{barbieri-etal-2020-tweeteval, meaney-etal-2021-semeval} to report the same evaluation metrics from the original tasks. 
Specifically, we report macro F1 for \underline{Emoji}, \underline{Hate}, \underline{Offensive}, and \underline{Humor}, report micro-Recall for \underline{Sentiment}, report the average of the F1 of \textit{against} and \textit{favor} (marked as F1$^{a+f}/2$) for \underline{Stance}, and report the F1 of the \textit{ironic} label (marked as F1$^i$) for \underline{Irony}.

\paragraph{Baselines and Comparisons.} 
For the \textbf{deduplication experiment}, we compare the proposed generative deduplication with two widely used deduplication approaches: \underline{shingles} \citep{broder1997resemblance} and \underline{simhash} \citep{manku2007detecting}. 
We also compare it with pairwise semantic deduplication using pretrained sentence embeddings \citep{li2023angle,li2024aoe}.
For \textbf{social media language understanding}, we adopt three popular backbones: \underline{RoBERTa} \citep{liu2019roberta}, \underline{BERTweet} \citep{nguyen-etal-2020-bertweet}, and \underline{LLaMA} \citep{touvron2023llama} and compare with them using deduplicated data.

\paragraph{Model Settings.} 
For generative deduplication, we use the T5-base model as the default generative backbone with a learning rate of $1e^{-4}$, beam size of $1$, and a prediction threshold of $0.5$.
For social media language understanding tasks, we use RoBERTa-base, BERTweet-base, and LLaMA-7B as backbones. 
For efficient training, we employ the LoRA \citep{hu2021lora} technique for LLaMA-7B fine-tuning with specific parameters ($lora\_r=16$, $lora\_alpha=16$, and $lora\_dropout=0.1$). 
The batch size is chosen from values $\{16, 32, 64, 128\}$ on the validation data. The initial learning rate is $2e^{-5}$ for BERT/RoBERTa-based models and $2e^{-4}$ for LLaMA-based models.


\paragraph{Dataset Deduplication.} 
Here, we first employ various deduplication approaches, including shingles, pairwise semantic deduplication, and the proposed generative deduplication, to eliminate duplicate training samples from social media language understanding datasets. 
Table \ref{table_dataset_dedup} shows the deduplicated training data size and time consumption of each dataset.
We can see that the proposed generative deduplication is more effective at removing redundant data than baselines with fewer training samples after deduplication. 
Also, the generative deduplication is efficient. 
It achieves competitive deduplication time with shingles and is highly efficient compared to pairwise deduplication.
Moreover, we perform random deduplication for a comprehensive evaluation. 
We randomly reduce the dataset size to match that of the generative deduplication. 
These deduplicated datasets by different approaches will be used for social media language understanding in Section \ref{sec::main-experiment}. 
Note that we only deduplicate the train set for model training, while keeping the validation and test sets for evaluation.

\input{table_dataset_dedup}

%% file: table_dataset_statistics.tex
\begin{table}[]
    \small
    \centering
    \begin{tabular}{l|rrr|r}
    \toprule
    \textbf{Dataset} & \textbf{Train} & \textbf{Valid.} & \textbf{Test} & \textbf{Labels}\\
    \midrule
    Emoji & $45,000$ & $5,000$ & $50,000$ & $20$ \\
    Hate & $9,000$ & $1,000$ & $2,970$ & $2$ \\
    Offensive & $11,916$ & $1,324$ & $860$ & $2$ \\
    Sentiment & $45,389$ & $2,000$ & $11,906$ & $3$\\
    Stance & $2,620$ & $294$ & $1249$ & $3$ \\
    Irony & $2,862$ & $955$ & $784$ & $2$ \\
    Humor & $8,000$ & $1,000$ & $1,000$ & $2$ \\
    \bottomrule
    \end{tabular}
    \caption{Number of labels and instances in training, validation (\textbf{valid.}), and test sets for the adopted datasets.}
    \label{table_dataset_statistics}
\end{table}

%% file: table_dataset_dedup.tex
\begin{table}[]
    \setlength\tabcolsep{5pt}
    \centering
    \small
    \begin{tabular}{lrrrr}
        \toprule
        \textbf{Dataset} & \textbf{Raw} & \textbf{SD} & \textbf{PD} & \textbf{GD} \\
        \midrule
        \multirow{2}{*}{Sentiment} & $45,615$ & $43,951$ & $35,605$ & $22,418$ \\
        & \textit{time} $\rightarrow$ & $1,074$ & $29,212$ & $808$ \\
        \midrule
        \multirow{2}{*}{Emoji} & $45,000$ & $40,656$ & $35,170$ & $31,425$ \\
        & \textit{time} $\rightarrow$ & $1,043$ & $29,047$ & $901$ \\
        \midrule
        \multirow{2}{*}{Offensive} & $11,916$ & $11,013$ & $10,771$ & $9,595$ \\
        & \textit{time} $\rightarrow$ & $78$ & $2,099$ & $240$ \\
        \midrule
        \multirow{2}{*}{Hate} & $9,000$ & $8,810$ & $7,818$ & $8,061$ \\
        & \textit{time} $\rightarrow$ & $50$ & $1,251$ & $178$ \\
        \midrule
        \multirow{2}{*}{Humor} & $8,000$ & $7,720$ & $7,848$ & $7,537$ \\
        & \textit{time} $\rightarrow$ & $39$ & $964$ & $153$ \\
        \midrule
        \multirow{2}{*}{Irony} & $2,862$ & $2,841$ & $2,712$ & $2,472$ \\
        & \textit{time} $\rightarrow$ & $9$ & $135$ & $59$ \\
        \midrule
        \multirow{2}{*}{Stance} & $2,620$ & $2,554$ & $2,547$ & $956$ \\  
        & \textit{time} $\rightarrow$ & $11$ & $113$ & $57$ \\  
        \midrule
        \multirow{2}{*}{\textbf{Total}} & $125,013$ & $117,545$ & $102,471$ & $82,464$ \\  
        & \textit{time} $\rightarrow$ & $2,304$ & $62,771$ & $2,396$ \\  
        \bottomrule
    \end{tabular}
    \caption{Deduplicated training data size and deduplication time (in seconds) of different approaches on different datasets. Raw is the original size. SD means Shingles Deduplication. PD stands for Pairwise Deduplication. GD is the proposed generative deduplication}
    \label{table_dataset_dedup}
\end{table}

%% file: main_experiment.tex

\subsection{Deduplication Results}
\label{sec-dedup-result}
The deduplication results are presented in Table \ref{table_dedup}. 
Our proposed generative deduplication outperforms the baselines.
Notably, the T5-base generative deduplication achieves a $23.4\%$ improvement in F1$^{eq}$ compared to shingles. 
Similarly, the T5-small generative deduplication shows a $19.8\%$ F1$^{eq}$ gain and reduces the deduplication time by $6.3$ seconds compared to shingles.
These improvements are attributed to the ability of generative deduplication to understand and remove semantic duplicates. 
In contrast, shingles and simhash, which focus on surface linguistic features, struggle with semantic-level duplication.
Furthermore, the proposed generative deduplication outperforms pairwise semantic deduplication and is more efficient. For instance, the T5-base generative deduplication is faster, taking $18.8$ seconds compared to $43.7$ seconds for pairwise semantic deduplication.

\input{table_dedup}

\input{table_main}

\input{figure_training_time_bar}

\subsection{Main Results}
\label{sec-main-result}
We show the main experimental results of social media language understanding tasks in Table \ref{table_main} and draw the following observations.

First, we can see that LLaMA-based models outperform RoBERTa and BERTweet-based models. 
This can be attributed to its larger model scale and powerful language understanding capability.
Second, we can find that the performance using random deduplication (RD) data is poorer than using other deduplication data. It negatively impacts performance. 
This might be because it eliminates non-duplicate patterns randomly and fails to address redundancy bias.
Third, it can be seen that shingles deduplication (SD) and pairwise semantic deduplication (PD) slightly improve the performance. 
This could be attributed to their ability to remove duplicate data partially and mitigate redundancy bias.
Fourth, our proposed generative deduplication (GD) consistently outperforms baselines trained on raw and other deduplicated training sets in average scores. 
Notably, generative deduplication generally performs better than shingles and pairwise semantic deduplication in various tasks, except for Stance detection using BERTweet. 
We will explain this exception in Section \ref{discuss-redundancy-bias}. 
Generative deduplication outperforms shingles because it can identify and remove semantically duplicate data, effectively reducing redundancy bias. 
Also, generative deduplication is more effective than pairwise semantic deduplication due to its stronger language understanding capabilities enabled by generative models. 
It can better identify and remove semantically duplicate data, further mitigating redundancy bias.

Furthermore, we compare the training time using different deduplicated datasets in Figure \ref{figure_training_time_bar}.
The generative deduplication can significantly reduce training time. This highlights generative deduplication's superior efficiency.

\subsection{Ablation Study}
\label{sec-ablation}

We have demonstrated the overall effectiveness of generative deduplication in the main results. 
Here, we further test its different settings via the ablation study. The results are presented in Table \ref{table_ablation}.

T5-base generative deduplication outperforms T5-small and can reduce more training samples.
It is attributed to T5-base's more powerful language understanding capabilities (with a larger model size), allowing it to handle noisy social media data more effectively. 

The results of generative deduplication with and without TGN indicate that TGN can improve generative deduplication performance.
This highlights that TGN effectively limits keyword prediction for non-duplicate texts by adding training difficulties and preventing trivial feature learning.

Finally, the results of one- and two-epoch training show that one-epoch outperforms two-epoch training. 
This is because generative models can replay keyword predictions for non-duplicate texts after multiple training epochs, resulting in the misidentification of non-duplicate texts.

\input{table_ablation}

\subsection{General Short Text Classification}
\label{sec-general}
We have shown that our generative deduplication method effectively challenges social media language understanding tasks. To further demonstrate the generality of our approach, we also evaluate generative deduplication on three general short-text classification tasks, including AGNews \citep{zhang2015character}, Subj \citep{pang-lee-2004-sentimental}, and SST-2 \citep{socher-etal-2013-recursive}. 
The results are presented in Table \ref{table_general}. The proposed generative deduplication can reduce the training data size, while keeping even improving the model performance slightly. 
Reduced training data size and improved performance demonstrate that the proposed generative deduplication effectively mitigates redundancy bias and can benefit general scenarios.

\input{table_general}

\subsection{Further Discussions}
\label{sec-discussion}

\paragraph{Discussion of Self-supervised Task.}
\label{discuss-self-sup-task}
In previous experiments, we have proven the effectiveness of the self-supervised keyword prediction task.
Here, we examine other self-supervised tasks to provide further insights. 
The results are presented in Table \ref{table_self_sup}.
The table shows that predicting a text's first or last word significantly reduces the training size. 
However, they yield poorer performance than others, possibly because they remove too many non-duplicate patterns. 
In contrast, we can observe that the random word prediction task has minimal impact on reducing the training size because it is more challenging for the model to learn.
Notably, single keyword (obtained by KeyBERT \citep{grootendorst2020keybert}) and multiple keywords (obtained by ChatGPT \citep{kim-etal-2023-chatgpt}) prediction outperform the other self-supervised tasks. 
It is because keywords convey the main idea of a text and enhance semantic learning, thus improving the text understanding and deduplication performance. 
We choose the KeyBERT-extracted keyword for self-supervised learning by default since it is more efficient and cheaper than ChatGPT-generated keywords and achieves similar performance as ChatGPT.

\paragraph{Discussion of Redundancy Bias.}
\label{discuss-redundancy-bias}
\input{figure_redundancy_bias}

To illustrate the redundancy bias intuitively, we present a plot of prediction confidence in Figure \ref{figure_redundancy_bias}. 

For the top 4 plots, we can see that the prediction confidence distribution of duplicate texts shifts towards a higher confidence zone than raw texts, suggesting a possible bias in the model.
The bias can lead to incorrect predictions for input text.
For example, in Figure \ref{figure_example}, the model has incorrect prediction biased by the common ``Pumpkin''  features with duplicate texts. 
Results in Table \ref{table_main} can also support this claim. The models trained on the generative deduplication data (less redundancy bias) outperform those without in these 4 tasks, except for LLaMA in Emoji prediction.

In contrast, the 3 bottom plots do not display notable distribution deviations, indicating that redundancy bias is not prominent in these tasks. In such cases, deduplication may negatively impact performance.
This explains why the performance using generative deduplication data on BERTweet is lower than the raw training data for Humor and Stance detection tasks in Table \ref{table_main}.

\input{table_self_sup}

\paragraph{Dicussion of Generative Deduplication Quality.} 
Figure \ref{figure_sim_dist} shows the pairwise similarity distribution for duplicate and non-duplicate texts (identified through generative deduplication). 
It is expected that duplicate texts would display higher similarity than non-duplicate texts. 
As observed, the similarity distribution of duplicate texts is shifted towards higher similarity values, indicating the good quality of generative deduplication.
We also use LLM Claude Sonnet 3.5 to evaluate the quality of the generative deduplication. About $66\%$ of the samples from the Irony dataset were considered semantically similar by Claude Sonnet 3.5. Table \ref{table_gd_random_samples} shows 10 random samples of generative deduplication.

\input{figure_tgn_sim}

\input{table_gd_random_samples}

\paragraph{Effect of TGN.} 
In Section \ref{sec-ablation}, we have shown the effectiveness of the proposed TGN mechanism. 
Here, we further discuss it by comparing it to similar existing mechanisms.

First, TGN is analogous to the mask mechanism used in masked language models such as BERT \citep{devlin-2019-bert}. 
The key difference is that we employ Gaussian noise instead of a fixed special mask token. TGN is more difficult than the mask mechanism. 
This design effectively limits the language understanding capabilities, preventing learning trivial features during generative training. As a result, it restricts keyword prediction for non-duplicate texts, thereby reducing misidentification.

Second, TGN also differs from the Dropout \citep{srivastava2014dropout} mechanism. 
Dropout is commonly applied to the feature dimension, not the time dimension, and can potentially improve language understanding capabilities by addressing overfitting.
Our experiment on the Emoji prediction demonstrates that replacing the proposed TGN with the Dropout mechanism leads to an inferior performance of $28.9$ compared to TGN's $31.4$.

Third, Figure \ref{figure_tgn} shows the TGN mechanism limits language understanding capabilities, as shown the model without TGN has a higher generation confidence than with.

%% file: table_dedup.tex
\begin{table}[]
    \centering
    \small
    \begin{tabular}{lcc}
        \toprule
        \textbf{Model} & \textbf{F1}$^{eq} \uparrow$ & \textbf{Time} (s) $\downarrow$ \\
        \midrule
        shingles & $32.9 \pm 0.0$ & $11.9$ \\
        simhash & $24.9 \pm 0.0$ & $\mathbf{1.4}$ \\
        pairwise semantic dedup. & $51.7 \pm 0.0$ & $42.7$ \\
        \midrule
        Generative Dedup. (T5-small) & $52.7 \pm 0.3$ & $5.6$ \\
        Generative Dedup. (T5-base) & $\mathbf{56.3} \pm 0.2$ & $18.8$ \\
        \bottomrule
    \end{tabular}
    \caption{Deduplication performance on MRPC dataset. Bold indicates the best results. $\uparrow$ means the higher, the better. $\downarrow$ stands for the smaller, the higher.}
    \label{table_dedup}
\end{table}

%% file: table_main.tex
\begin{table*}[ht]
\small
\setlength\tabcolsep{2.5pt}
\centering
\begin{threeparttable}
\begin{tabular}{lcccccccc}
\toprule
        \multirow{2}{*}{\textbf{Model}}   & \multicolumn{1}{c}{\textbf{Emoji}} & 
        \multicolumn{1}{c}{\textbf{Hate}} & \multicolumn{1}{c}{\textbf{Offensive}} & \multicolumn{1}{c}{\textbf{Humor}} & \multicolumn{1}{c}{\textbf{Sentiment}} &  \multicolumn{1}{c}{\textbf{Stance}} &  \multicolumn{1}{c}{\textbf{Irony}} & \multirow{2}{*}{\textbf{Avg.}}\\

        \cmidrule{2-5} \cmidrule{6-8} &   &  \multicolumn{2}{c}{\textbf{Macro F1} $\uparrow$} & & \multicolumn{1}{c}{\textbf{Macro Recall} $\uparrow$} & \multicolumn{1}{c}{\textbf{F1}$^{a+f}/2 \uparrow$} &  \multicolumn{1}{c}{\textbf{F1}$^{i} \uparrow$}\\

\midrule
\textbf{RoBERTa} \\
\ Raw & $30.9 \pm 0.2 \ \diamondsuit$ & $46.6 \pm 1.8 \ \diamondsuit$ & $79.5 \pm 0.7\ \diamondsuit$ & \cellcolor{lightblue}$95.0 \pm 0.6 \ \dagger$ & $71.3 \pm 1.1 \ \diamondsuit$ & $68.0 \pm 0.8 \ \diamondsuit$ & $59.7 \pm 5.0 \ \diamondsuit$ & $64.4$ \\
\ RD & $29.8 \pm 0.3$ & $45.7 \pm 1.6$ & $79.0 \pm 0.9 $ & $93.3 \pm 0.5$ & $71.5 \pm 1.3$ & $65.1 \pm 1.2$ & $58.5 \pm 2.1$ & $63.3$ \\
\ SD & $31.0 \pm 0.3$ & $47.3 \pm 1.3$ & $79.4 \pm 0.6$ & $93.6 \pm 0.5$ & $71.5 \pm 1.3$ & $67.9 \pm 0.9$ & $61.3 \pm 3.7$ & $64.6$ \\
\ PD & $31.1 \pm 0.3$ & $47.7 \pm 1.3$ &  $79.7 \pm 0.7$ &  $94.0 \pm 0.3$ &  $71.5 \pm 1.1$ & $68.1 \pm 0.8$ & $62.0 \pm 2.7$ & $64.9$ \\
\ GD & \cellcolor{lightblue}$31.4 \pm 0.2$ & \cellcolor{lightblue}$49.5 \pm 1.1$ & \cellcolor{lightblue}$80.7 \pm 0.7$ & $94.3 \pm 0.4$ & \cellcolor{lightblue}$71.8 \pm 1.1$ & \cellcolor{lightblue}$68.3 \pm 0.6$ & \cellcolor{lightblue}$62.6 \pm 1.8$ & \cellcolor{lightblue}$65.5$ \\

\midrule
\textbf{BERTweet} \\
\ Raw & $32.3 \pm 0.5$ & $54.9 \pm 0.9\ \dagger$ & $80.5 \pm 0.8 \ \dagger$ & \cellcolor{lightblue}$\mathbf{95.9} \pm 0.3\ \dagger$ & \cellcolor{lightblue}$72.3 \pm 1.2$ & \cellcolor{lightblue}$70.3 \pm 0.9 \ \dagger$ & $78.7 \pm 1.4 \ \dagger$ & $69.3$ \\
\ RD & $31.2 \pm 0.7$ & $54.5 \pm 0.9$ & $80.2 \pm 1.0 $ & $94.5 \pm 0.4$ & $71.4 \pm 1.7$ & $65.9 \pm 1.3$ & $77.9 \pm 1.5$ & $67.9$ \\
\ SD & $32.4 \pm 0.5$ & $55.0 \pm 0.8$ & $80.5 \pm 0.8$ & $94.5 \pm 0.3$ & $72.1 \pm 1.1$ & $69.9 \pm 1.1$ & $78.7 \pm 1.4$ & $69.0$ \\
\ PD & $32.5 \pm 0.4$ & $55.3 \pm 1.0$ &  $80.6 \pm 1.0$ &  $94.5 \pm 0.5$ &  $72.2 \pm 1.3$ & $69.4 \pm 1.0$ & $79.1 \pm 1.5$ & $69.1$ \\
\ GD & \cellcolor{lightblue}$32.6 \pm 0.3$ & \cellcolor{lightblue}$55.7 \pm 1.0$ & \cellcolor{lightblue}$80.9 \pm 0.6$ & $95.0 \pm 0.3$ & $72.2 \pm 1.0$ & $69.3 \pm 0.8$ & \cellcolor{lightblue}$\mathbf{80.1} \pm 1.2$ & \cellcolor{lightblue}$69.4$ \\

\midrule
\textbf{LLaMA} \\
\ Raw & \cellcolor{lightblue}$\mathbf{37.4} \pm 0.6$ & $58.2 \pm 1.3$ & $80.7 \pm 1.2$ & \cellcolor{lightblue}$95.3 \pm 0.5$ & $73.0 \pm 1.2$ & $70.1 \pm 0.7$ & $74.5 \pm 1.1$ & $69.9$ \\
\ RD & $36.2 \pm 1.1$ & $57.7 \pm 1.8$ & $79.8 \pm 1.3$ & $95.1 \pm 0.8$ & $71.6 \pm 1.6$ & $66.5 \pm 0.9$ & $73.2 \pm 1.6$ & $68.6$ \\
\ SD & $37.1 \pm 0.9$ & $58.1 \pm 1.3$ & $80.3 \pm 1.4$ & $95.2 \pm 0.6$ & $73.1 \pm 1.3$ & $70.4 \pm 0.6$ & $75.6 \pm 1.3$ & $70.0$ \\
\ PD & $37.3 \pm 1.1 $ & $58.3 \pm 1.4$ & $80.8 \pm 1.3$ & $95.2 \pm 0.6$ & $73.0 \pm 1.4$ & $70.8 \pm 0.8$ & $75.9 \pm 1.5$  & $70.2$\\
\ GD & $37.3 \pm 0.8$ & \cellcolor{lightblue}$\mathbf{58.6} \pm 1.4$ & \cellcolor{lightblue}$\mathbf{81.0} \pm 1.3$ & \cellcolor{lightblue}$95.3 \pm 0.6$ & \cellcolor{lightblue}$\mathbf{73.5} \pm 0.9$ & \cellcolor{lightblue}$\mathbf{71.1} \pm 1.1$ & \cellcolor{lightblue}$76.4 \pm 1.2$ & \cellcolor{lightblue}$\mathbf{70.5}$ \\
\bottomrule
\end{tabular}
\end{threeparttable}
\caption{Results of social media language understanding tasks. $\diamondsuit$: results are from \citet{barbieri-etal-2020-tweeteval}. $\dagger$: results are retrieved from \citet{tan2023hicl}. 
We follow previous work to report the average result of five runs. 
``Raw'' refers to the use of the original train set. 
``RD'', ``SD'', ``PD'', and ``GD'' are trained on the deduplicated training set of random, shingles, pairwise semantic, and generative deduplication, respectively. The light blue color indicates the best results for each backbone, while the bold marks the best overall results.
}
\label{table_main}
\end{table*}

%% file: figure_training_time_bar.tex
\begin{figure}[ht]
    \centering
    \includegraphics[width=0.98\textwidth]{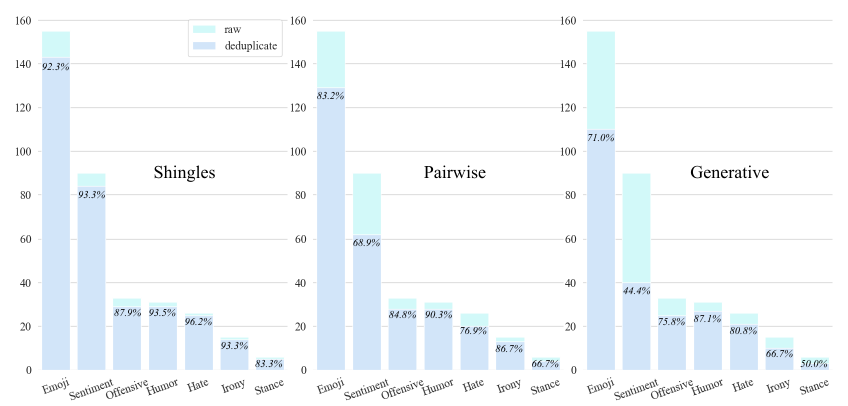}
    \caption{A bar plot shows the training time of RoBERTa-based models for each dataset. The percentage represents the training time on deduplicated data compared to the training time on raw data.}
    \label{figure_training_time_bar}
\end{figure}

%% file: table_ablation.tex
\begin{table}[]
    \centering
    \small
    \setlength\tabcolsep{2.5pt}
    \begin{tabular}{lcc}
        \toprule
        \textbf{Model} & \textbf{Train Size} & \textbf{Macro F1} $\uparrow$ \\
        \midrule
        \textbf{Generative Dedup. (T5-base)} & $31.4K$ & $\mathbf{31.4}$ \\
        \ w/o TGN & $30.6K$ & $30.1$\\
        \ epoch=2 & $29.7K$  & $29.8$ \\
        \midrule
        \textbf{Generative Dedup. (T5-small)} & $36.2K$ & $30.0$ \\
        \ w/o TGN & $34.6K$ & $28.7$ \\
        \ epoch=2 & $34.4K$ & $28.3$ \\
        \bottomrule
    \end{tabular}
    \caption{Ablation study of generative deduplication on the \textbf{Emoji} prediction task. $K$ represents thousands. The train size of the raw \textbf{Emoji} is $45K$. $K$ stands for thousand. Bold indicates the best results.}
    \label{table_ablation}
\end{table}

%% file: table_general.tex
\begin{table}[]
    \setlength\tabcolsep{2.5pt}
    \centering
    \small
    \begin{tabular}{l|cc|cc|cc}
        \toprule
        \multirow{2}{*}{\textbf{Model}} & \multicolumn{2}{c|}{\textbf{AGNews}} & \multicolumn{2}{c|}{\textbf{Subj}} & \multicolumn{2}{c}{\textbf{SST-2}} \\
        & data $\downarrow$ & acc $\uparrow$ & data $\downarrow$ & acc $\uparrow$ & data $\downarrow$ & acc $\uparrow$ \\
        \midrule

        \textbf{RoBERTa} & $120000$ & $94.88$ & $8000$ & $96.60$ & $6920$ & $94.29$  \\
        \ w/ RD & $47786$ & $94.31$ & $7127$ & $96.65$ & $6276$ & $94.23$ \\
        \ w/ SD & $104539$ & $\mathbf{94.91}$ & $7925$ & $97.00$ & $6690$ & $94.45$ \\
        \ w/ PD & $77789$ & $94.80$ & $7930$ & $97.15$ & $6689$ & $95.00$ \\
        \ w/ GD & $47786$ & $94.83$ & $7127$ & $\mathbf{97.30}$ & $6276$ & $\mathbf{95.05}$ \\
        \bottomrule
    \end{tabular}
    \caption{Results on general short-text classification. data represents the training data size. acc denotes the evaluation metric Accuracy (\%). $\downarrow$ denotes the smaller the better, while $\uparrow$ means the larger the better.}
    \label{table_general}
\end{table}

%% file: figure_redundancy_bias.tex
\begin{figure*}[ht]
    \centering
    \includegraphics[width=0.98\textwidth]{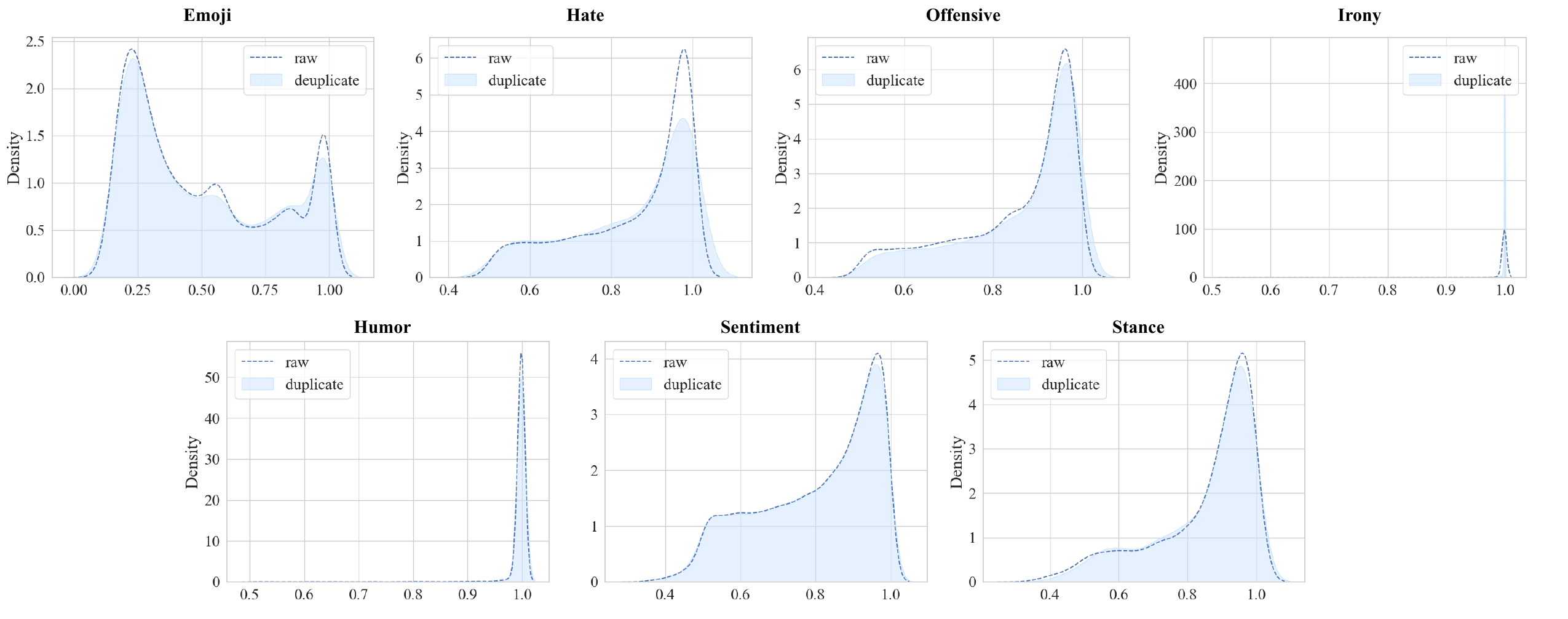}
    \caption{The Kernel Density Estimate (KDE) plot of the prediction confidence of RoBERTa-based models on the training set for each task. $x$-axis indicates the confidence. The top 4 plots present bias in the high confidence zone. The bottom 3 plots do not show obvious bias.}
    \label{figure_redundancy_bias}
\end{figure*}

%% file: table_self_sup.tex
\begin{table}[]
    \setlength\tabcolsep{3pt}
    \centering
    \small
    \begin{tabular}{lcc}
        \toprule
        \textbf{Model} & \textbf{Train Size} & \textbf{Macro F1} $\uparrow$ \\
        \midrule
        single keyword (KeyBERT) & $31.4K$ & $31.4$ \\
        multiple keywords (ChatGPT) & $37.6K$ & $\mathbf{31.8}$ \\
        first word & $11.4K$ & $21.6$\\
        last word & $15.1K$  & $24.5$ \\
        middle word & $26.6K$  & $28.9$ \\
        random word & $44.9K$  & $30.6$ \\
        \bottomrule
    \end{tabular}
    \caption{The results of different generative self-supervised tasks on the \textbf{Emoji} prediction task. $K$ represents thousands. The train size of the raw \textbf{Emoji} is $45K$. T5-base is the generative backbone for generative deduplication, and RoBERTa-base serves as the downstream backbone model. Bold indicates the best results.}
    \label{table_self_sup}
\end{table}

%% file: figure_tgn_sim.tex
\begin{figure}
     \centering
     \begin{subfigure}[b]{0.49\textwidth}
         \centering
         \includegraphics[width=1.0\textwidth, trim = {10 10 10 10}]{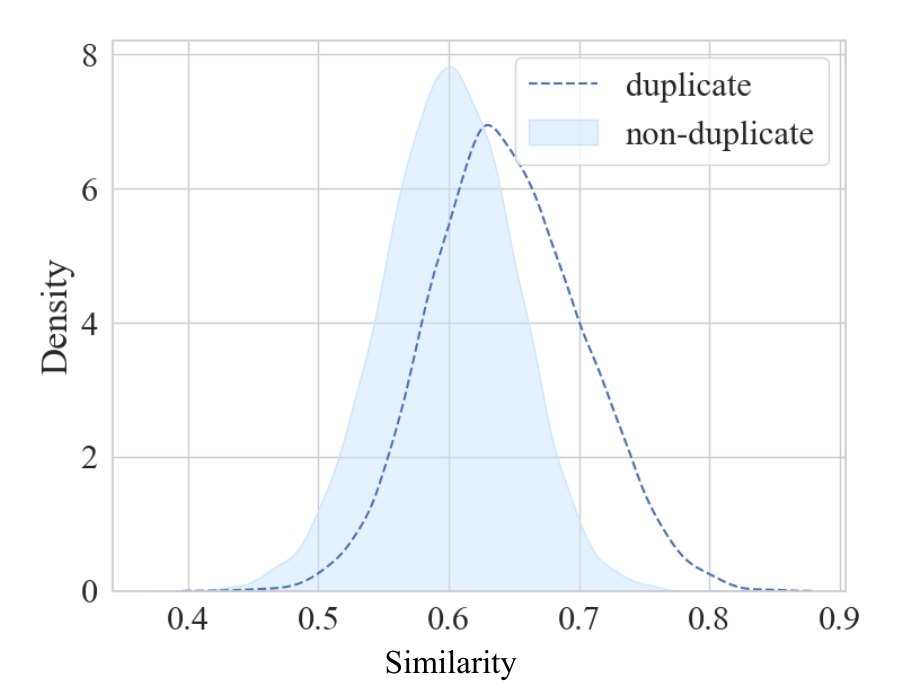}
         \caption{}
         \label{figure_sim_dist}
     \end{subfigure}
     \begin{subfigure}[b]{0.49\textwidth}
         \centering
         \includegraphics[width=1.0\textwidth, trim = {10 10 10 10}]{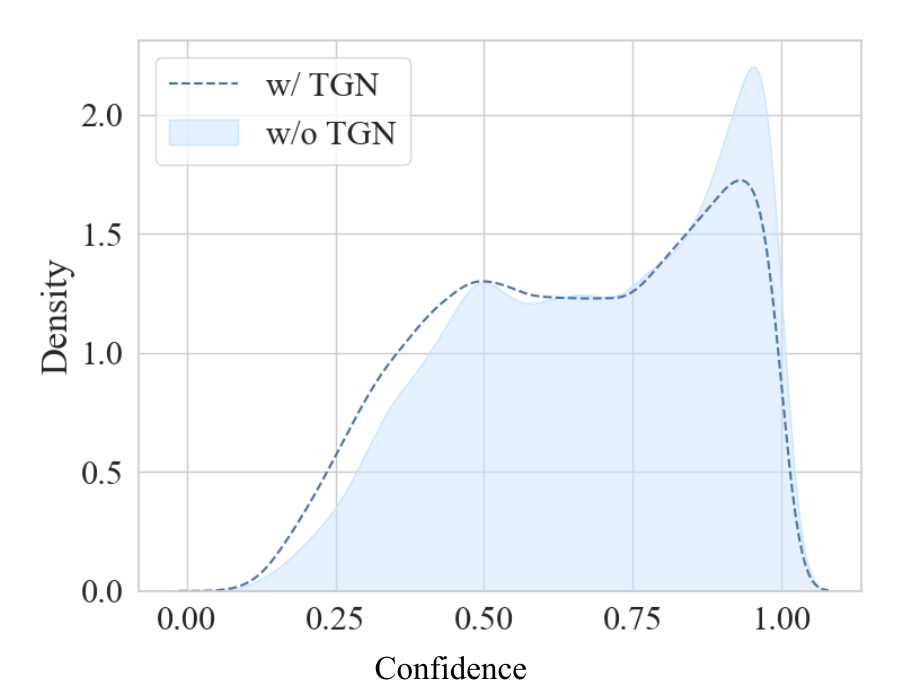}
         \caption{}
         \label{figure_tgn}
     \end{subfigure}     
     \caption{(a) The similarity distribution for duplicate and non-duplicate texts, which are identified by generative deduplication, for all datasets. (b) The KDE plot of the generation probability (confidence) with and without TGN on \textbf{Emoji} generative deduplication.}
\end{figure}

%% file: table_gd_random_samples.tex
\begin{table*}[t]
\centering
\small
\begin{tabular}{p{0.42\textwidth} p{0.42\textwidth} c}
\toprule
\textbf{Text 1} & \textbf{Text 2} & \textbf{Label} \\
\midrule
Today is awesome & Today is awesome! & 1 \\
\midrule
I have such a loving family & Dead supportive family I've got. & 0 \\
\midrule
At least I woke up feeling a lot better today.. & Yeah this is so good just woke up & 1 \\
\midrule
I absolutely LOVE moving house & Dayum, I really got the house to myself while my brother still has school all week & 0 \\
\midrule
Well, weekend is over!|Now it's \#TwitterTime again :D|Have a nice monday!| & Nice weekend off but back at work tonight. & 1 \\
\midrule
working on my birthday \#yay \#sucks & Love the fact I'm sick on my birthday & 1 \\
\midrule
Be Blessed friends. Merry Christmas to all! & Merry Christmas @user & 1 \\
\midrule
Great start to the day & Great way to start of the day & 1 \\
\midrule
Love these cold winter mornings  best feeling everrrrrrr! & I love cold winter days cause I never know when my car decides not to start & 0 \\
\midrule
It's 8:46 and I'm ready for bed. & I am now heading for bed orz & 1 \\
\bottomrule
\end{tabular}
\caption{Ten random samples of generative deduplication from the Irony dataset. The label indicates the duplication judgment by Claude Sonnet 3.5. A label of 1 means that text 1 and text 2 are duplicates, 0 is non-duplicate.}
\label{table_gd_random_samples}
\end{table*}

%% file: acknowledgement.tex
This work is supported by a grant from the Research Grants Council of the Hong Kong Special Administrative Region, the NSFC Young Scientists Fund (Project No. 62006203), China (Project No. PolyU/25200821), the Innovation and Technology Fund (Project No. PRP/047/22FX), PolyU Internal Fund from RC-DSAI (Project No. 1-CE1E), and PolyU Embodied Artificial Intelligence Lab (No. N-ZGNN).

Other than that, we sincerely thank the reviewers and ACs for their valuable input, which has greatly improved our work.